# Robotics Technology in Mental Health Care

## Laurel D. Riek

Department of Computer Science and Engineering, University of Notre Dame

## INTRODUCTION

This chapter discusses the existing use of and future potential use of robotics technology in mental health care. Robotics technology primarily refers to robots, which are physically embodied systems capable of enact- ing physical change in the world. Robots enact this change with effectors that either move the robot itself (locomotion), or move objects in the environment (manipulation), and often use data from sensors to make decisions.

Robots can have varying degrees of autonomy, ranging from fully teleoperated (the operator makes all decisions for the robot) to fully autonomous (the robot is entirely independent). The term robotics technology also broadly includes affiliated technology, such as accompanying sensor systems, algorithms for processing data, etc.

As a discipline, robotics has traditionally been defined as "the science which studies the intelligent connections between perception and actions," though in recent years this has shifted outward, becoming focused on problems related to interacting with real people in the real world (Siciliano & Khatib, 2008). This shift has been referred to in the literature as human-centered robotics, and an emerging area in the past decade focusing on problems in this space is known as human-robot interaction (HRI).

The use of robotics technology in mental health care is nascent, but represents a potentially useful tool in the professional's toolbox. Thus, the goal of this chapter is to provide a brief overview of the field, discuss the recent use of robotics technology in mental healthcare practice, explore some of the design issues and ethical issues of using robots in this space, and finally to explore the potential of emerging technology.

## Background

## Human-Robot Interaction

Goodrich and Schultz (2007) describe the HRI problem as seeking "to understand and shape the interactions between one or more humans and one or more robots." They decompose the problem into five principle attributes: (i) the level and behavior of a robot's autonomy, (ii) the nature of information exchange between human and robot, (iii) the structure of the human-robot team, (iv) how people and robots adapt and learn from one another, and (v) how the task shapes interaction.

All of these factors play a role in how a mental healthcare professional might consider the use of robotics technology in their practice. However, there are two additional factors that may be of particular importance to practitioners. The first is the morphology, or form, of the robot itself. Robots can range in appearance from very mechanical-looking to very anthropomorphic in appearance (Riek, Rabinowitch, Chakrabarti, & Robinson, 2009). Morphology is a richly debated topic in the research community, with many studies showing people will anthropomorphize and form attachments to nearly anything conveying animacy. Some researchers worry this not only conveys inaccurate expectations to people about a robot's capabilities, but may also be unethical when treating vulnerable populations (Riek, Hartzog, Howard, Moon, & Calo, 2015; Riek & Howard, 2014). For example, individuals with cognitive impairments or children may be more susceptible to deception and manipulation by robots.

A second factor that can impact the HRI problem is individual differences between people. People have a wide range of existing cognitive and physical attributes which can greatly influence how they perceive, interact with, and accept robots (Hayes & Riek, 2014). These factors may be particularly important when considering the use of robotics technology for clients in mental healthcare settings, who may have further unique needs.

Cite as: Riek, L.D. "Robotics Technology in Mental Healthcare". In D. Luxton (Ed.), *Artificial Intelligence in Behavioral Health and Mental Health Care*. Elsevier, 2015.

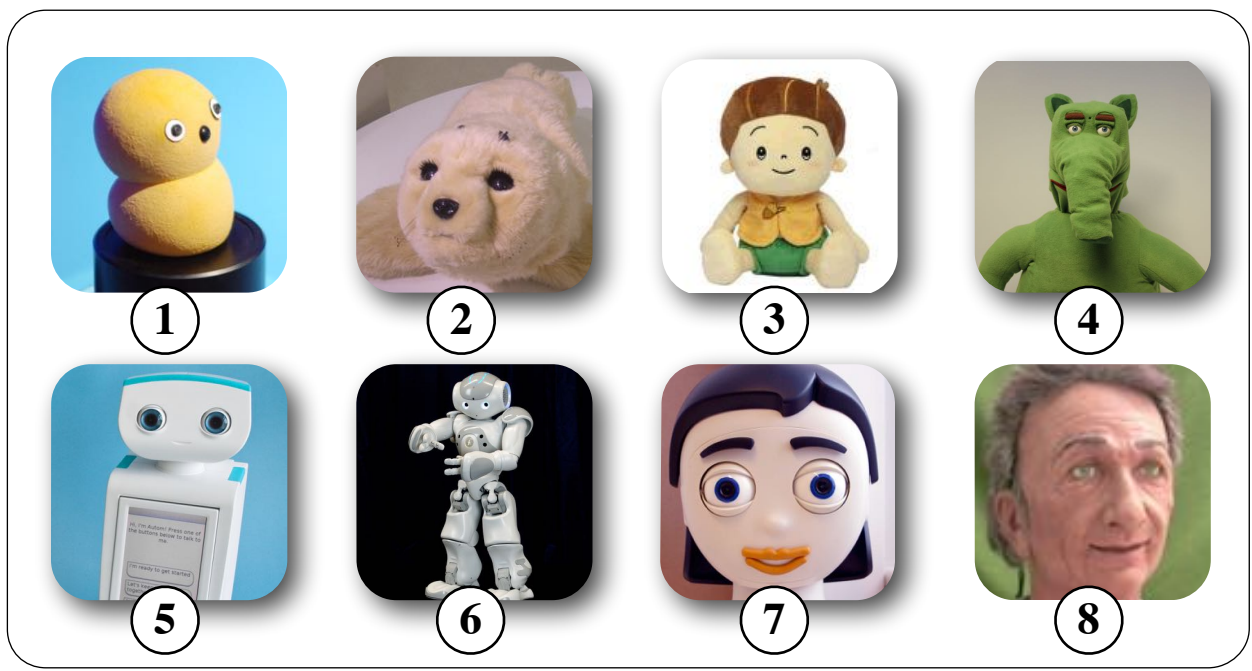

**Figure 8.1** Robots currently used in mental healthcare vary greatly in their morphology, and include zoomorphic, mechanistic, cartoon-like, and humanoid representations. These robots have been used for helping treat people with dementia, autism, and cognitive impairments; have helped provide companionship for people who were lonely; have been used to help educate children with developmental disabilities; and have been used to help improve how people with visible disabilities are treated. (1) KeepOn (Kozima, Michalowski, & Nakagawa, 2009), (2) Paro (Shibata, 2012), (3) Kabochan (Tanaka et al., 2012), (4) Probo (Vanderborght et al., 2012), (5) Autom (Kidd & Breazeal, 2008), (6) NAO (Diehl et al., 2014), (7) Flobi (Damm et al., 2013), and (8) Charles (Riek & Robinson, 2011).

## Robot Morphology

Figure 8.1 depicts several consumer and research robots used in mental healthcare applications, many of which will be discussed in next section. The robots in this figure are representative of the state-of the-art for socially interactive robots. Robots with more mechanistic appearances have been used in other applications, though very few in mental health care.

While zoomorphic, anthropomorphic, and cartoon-like morphologies are the most common, some robot designers have explored other unique representations. For example, actuating "everyday objects" like balls, drawers, and ottomans (Michaud, Duquette, & Nadeau, 2003; Sirikin & Ju, 2014). These robots are quite engaging, due to people's innate tendency to anthropomorphize anything with animacy (Heider & Simmel, 1944; Waytz, Cacioppo, & Epley, 2010). They may serve a useful role in therapeutic applications with clients who are less comfortable with anthropomorphic or zoomorphic representations, such as individuals on the autism spectrum (Diehl et al., 2014).

Often a robot's morphology is directly related to its functional capability requirements; for example, a robot that needs to manipulate objects is likely to have a grasper, and a robot that needs to climb stairs is likely to have legs. However, consumer robots often have appearances that reflect science fiction depictions in their color (grey) and shape (boxy). They also sometimes convey extreme feminine representations (i.e., fembots), which has raised ethical concerns in some research communities (Riek & Howard, 2014).

While consumers do not have much choice over the appearance of the robot they purchase, they frequently dress, name, and otherwise take steps to personalize it. For example, extensive, long-term, in-home studies of the Roomba vacuum cleaning robot reflect this consumer personalization (Forlizzi & DiSalvo, 2006; Sung, Guo, Grinter, & Christensen, 2007).

As will be discussed in the "Ethical Issues" section of this chapter, it is important that mental healthcare professionals are careful when selecting a robot morphology to use in treatment. Many people have a latent fear of robots due to 60 years of sordid science fiction depictions, and these fears could be exacerbated in a mental healthcare scenario. On the other hand, some morphologies may inhibit or delay transfer of learned skills from the therapy setting to everyday life. In general, when selecting a platform, mental healthcare professionals need to carefully weight the capabilities of the robot against the therapeutic needs for the patient.

**Robot Capabilities**

Current robots have an extensive range of physical capabilities, and as the service robotics industry continues to blossom these capabilities will only grow. In terms of physical capability, robots of various morphologies can exhibit limb-like motion, such as walking, running, climbing, turning, grabbing, shaking, gesturing; face-like motion, such as facial expressions, gaze, nodding; and other forms of biological motion, such as flipping, flying, and undulating.

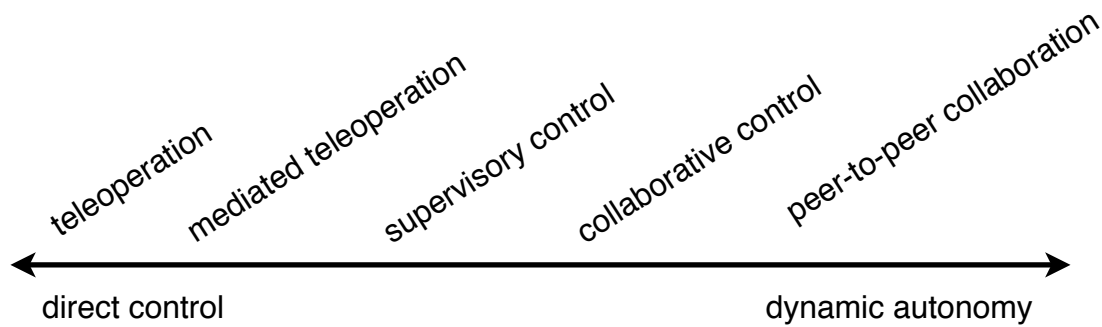

**Figure 8.2** The scale of autonomy which a robot may employ during an HRI scenario. Scale from Goodrich and Schultz (2007), used with permission.

However, the technology to enable these platforms to autonomously act safely and capably in the presence of humans is still very much nascent. While the research community has made significant strides in recent years, it still faces a number of challenges, particularly within domestic environments. Thus, the majority of present-day robots used in mental health care are either fully controlled by an operator (i.e., by a Wizard), or fully preprogrammed and thus somewhat limited in their abilities (Riek, 2012, 2013).

**Robot Autonomy**

Figure 8.2 depicts the types of autonomy a robot can employ in an HRI scenario. Many modern robotic systems have adjustable autonomy, where a person can change how they interact with a robot in real time (Dorais, Bonasso, Kortenkamp, Pell, & Schreckenghost, 1999). This is particularly useful in mental healthcare scenarios, where a professional may wish to have certain robot behaviors directly controlled and others autonomous. For example, a therapist working with a child with autism might want to have the robot autonomously play a game, and based on the child's prog- ress adjust how the robot administers rewards.

## Recent Applications

Robotics technology has been applied in a variety of ways in mental healthcare scenarios. Such applications include interventions for conditions ranging from autism spectrum disorder to cognitive impairments, as well as ways to encourage physical activity and provide companionship to individuals living alone.

It is worth noting that due to the nascency of consumer robotics technology and its tendency to change rapidly, few randomized controlled clinical trials (RCTs) have been reported in the literature. Many published studies have small sample sizes, lack adequate controls, and are difficult to replicate due to variability in hardware and software. Thus, one should be cautious when interpreting the efficacy of these interventions. That being said, there is still value in learning from qualitative and case- based studies, therefore some of these findings are also reported here.

### Autism Spectrum Disorders

One of the most common applications of robotics technology in mental health care is in the diagnosis and treatment of autism spectrum disorders (ASD). As Diehl et al. (2014) reported in a recent review article, robotics technology holds great potential for people with ASD, as they are very responsive to treatment involving robotics technology, possibly more so than treatment with human therapists. Thus, there are many recent studies in the literature reporting the use of robots in this fashion (cf. Damm et al., 2013; Feil-Seifer & Mataric´, 2009; Goodrich et al., 2012; Kozima et al., 2009; Robins & Dautenhahn, 2014; Scassellati, Admoni, & Mataric, 2012; Vanderborght et al., 2012).

However, Diehl et al. (2014) advise caution in deploying this technol- ogy clinically, as few RCTs have been conducted, and the majority of reported research in the literature has been more technology-focused rather than client-focused. Thus, the authors argue that using robots clin- ically for ASD diagnosis and treatment should be considered as an experi- mental approach, and suggest "clinical innovation should parallel technological innovation" if robots are to be fully realized in the ASD clinical space.

## Activity Engagement and Physical Exercise

Lancioni et al. describe several robot-based intervention studies intended to encourage activity engagement and ambulation among a small number of participants ($n = 6$) with severe physical and cognitive disabilities (Lancioni, Sigafoos, O'Reilly, & Singh, 2012). The mobile robots used in the studies helped participants' engagement with the physical world in a significant way, both increasing their independence as well as their "social image."

Numerous studies report the use of robotics technology in upper-limb therapy post-stroke (see Kwakkel, Kollen, and Krebs (2007) for a thorough review). However, more recently, a large-scale RCT ($n = 127$) conducted by Lo et al. (2010) showed no significant difference between the use of a robot vs. more conventional therapy. Furthermore, anecdotal evidence suggests that many of these upper-limb rehabilitative robots are so difficult for therapists to use that they remain dormant in closets after they are purchased.

In a different area of post-stroke rehabilitation, Mataric̓ et al. introduce the concept of socially assistive robots (SAR), which provide social or cognitive assistance to people without physically interacting with them.[1] SAR "focuses on achieving specific convalescence, rehabilitation, training, or education goals." Across several small ($n = 2$), non-RCT pilot studies, the authors have found positive effects from the use of a SAR for post-stroke activity encouragement (Mataric̓, Tapus, Winstein, & Eriksson, 2009).

Kidd and Breazeal report another application for using robots for encouraging activity engagement: weight loss. The authors designed a robotic weight loss coach, Autom, which after a controlled study ($n = 45$) was shown to be more effective at maintaining long-term weight loss (i.e., encouraging diet adherence and exercise) compared to a paper-based or computer-based system (Kidd & Breazeal, 2008).

---

[1] In later work, the authors extend this definition to include "robots that assist people with special needs through social interactions" (Scassellati et al., 2012).

## Dementia and Age-Related Cognitive Decline

Mordoch et al. provide a detailed review of the literature on robot use in dementia care, and report details from approximately 21 studies involving patients with dementia from 2004-2011 (Mordoch, Osterreicher, Guse, Roger, & Thompson, 2013). It should be noted that none of these studies included a randomized controlled trial, most have small sample sizes, the majority lack adequate controls, and most will likely be difficult to replicate due to a lack of standardized hardware/software (Broekens, Heerink, & Rosendal, 2009; Mordoch et al., 2013).

Shibata et al. reported numerous studies of patients with cognitive impairments who received Paro as a therapeutic invention (Shibata, 2012; Shibata & Wada, 2011). Paro is a zoomorphic platform that resembles a seal (see Figures 8.1 and 8.2), and has been used worldwide in care homes as a substitute for physical animal therapy. Therapy using Paro typically involves patients holding, hugging, stroking, or talking to Paro, as they would an actual animal or baby. The dementia-related studies reported by Shibata include a mix of short-term and long-term studies, with both qualitative and quantitative data collected. Therefore, it is difficult to draw any conclusions as to Paro's overall effectiveness in therapy for people with dementia.

For age-related cognitive decline, there was one well-designed RCT reported in the literature suggesting the effectiveness of a robot intervention. Tanaka et al. (2012) report a trial on 34 ($n = 34$) healthy Japanese women living alone, aged between 66 and 84 years old. The study lasted for 8 weeks. Participants in the intervention group interacted with the Kabochan Nodding Communication ROBOT, a cartoon-like platform that would talk and nod at participants (see Figure 8.1); while participants in the control group received a robot that had an identical morphology, but did not talk or nod. At the end of the study, participants in the intervention group had lower cortisol levels (as measured by their saliva), improved judgment and verbal memory function (as measured by the Mini-Mental State Examination), and improved nocturnal sleep (self-report).

**Companion Robots to Improve Psychosocial Outcomes**

Several controlled studies have suggested companion-like robots may be an effective treatment to reducing loneliness and lowering blood pressure. For example, a recent randomized controlled trial was conducted in 2013 by Robinson et al. at a care home in New Zealand ($n = 40$). Over the course of 12 weeks, participants in the intervention group interacted with Paro, and participants in the control group participated in standard activities at the care home. Participants who interacted with Paro showed a significant decrease in loneliness over the trial period compared to the control group (Robinson, MacDonald, Kerse, & Broadbent, 2013).

This effect of loneliness reduction from a pet-like robot has also been observed in other RCTs with AIBO, a robotic dog, in the US ($n = 25$). This effect was also seen in Japan across several between-subject and within-group experiments ($n = 13$, $n = 5$, respectively) (Banks, Willoughby, & Banks, 2008; Tamura et al., 2004). Bemelmans et al. present a detailed systematic review of this literature (Bemelmans, Gelderblom, Jonker, & de Witte, 2012).

In 2014, Robinson et al. published a pilot study in both a rest home and hospital setting with people briefly interacting with Paro in a repeated measures design ($n = 21$). Participants' blood pressure was taken before, during, and after interacting with the robot. The researchers found a significant decrease in systolic and diastolic blood pressure when participants had Paro, and a significant increase in diastolic blood pressure after the robot was withdrawn (Robinson, MacDonald, & Broadbent, 2014).

**Clinician Training for Interacting with People with Disabilities**

We have been researching novel ways to use humanoid robots to train clinicians to better interact with patients during face-to-face interaction (Gonzales, Moosaei, & Riek, 2013; Moosaei, Gonzales, & Riek, 2014; Riek & Robinson, 2011). This work was motivated by the fact that clinicians have been shown to express bias against people with both visible and invisible disabilities, which is thus a pertinent topic for mental health care (Deegan, 1990; Mason & Scior, 2004).

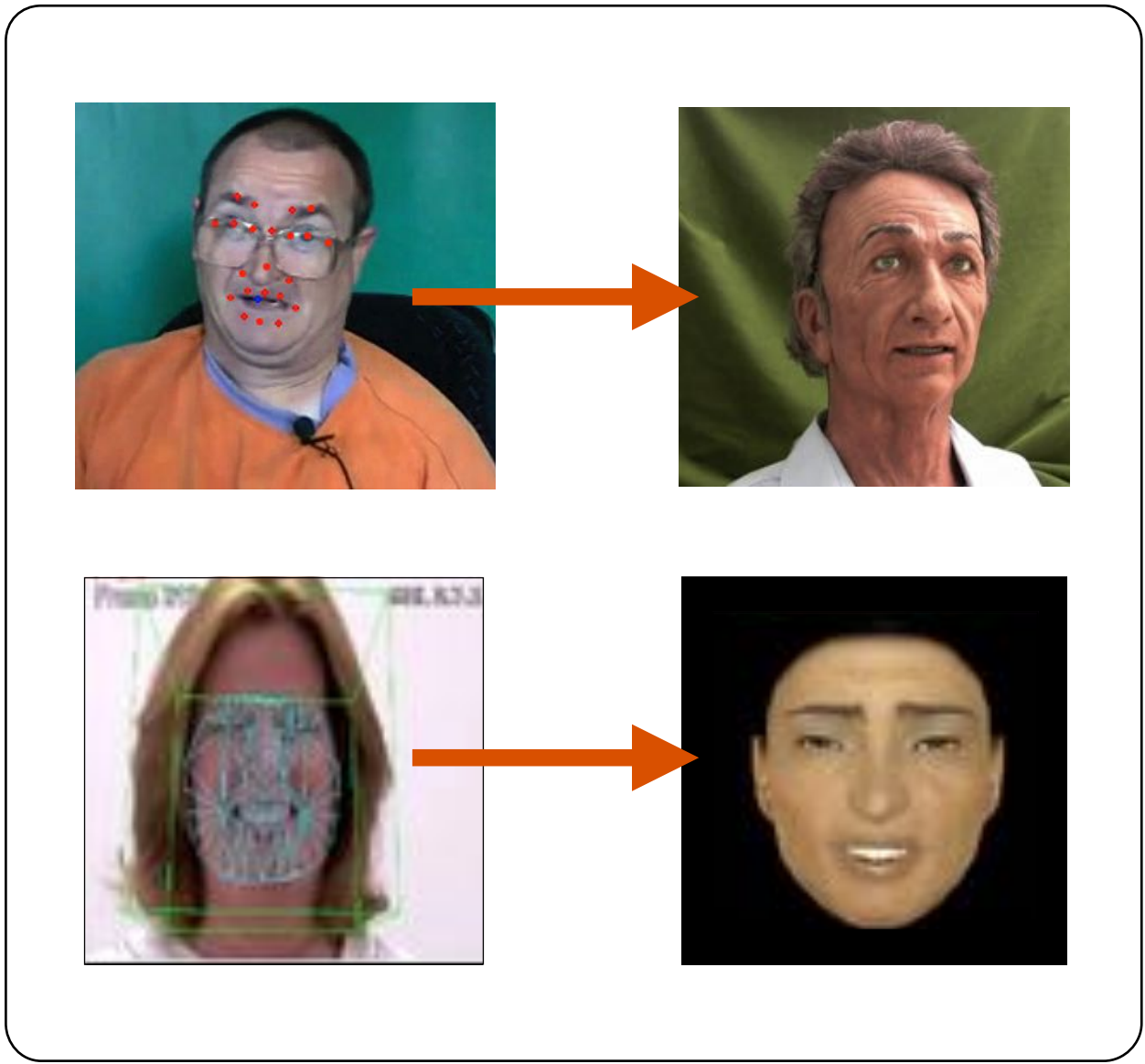

**Figure 8.3** Work by the author on naturalistic facial expression synthesis, based on real patients, to be used to train clinicians to better interact with people with disabilities. On the top, facial movements tracked on a person with cerebral palsy is used to move the robot Charles. On the bottom, facial movement from a person with chronic shoulder pain is used to animate a virtual patient (the author is currently transitioning this technology to a physical robot) (Moosaei et al., 2014; Riek & Robinson, 2011).

We have pioneered the concept of naturalistic, patient-based facial expression synthesis for simulated patients, and have synthesized pathologies including cerebral palsy, dystonia, and pain (see Figure 8.3). Mental healthcare training already includes the use of medical simulation technology, including life-sized human-patient simulators, which are essentially humanoid robots. We plan to conduct an RCT with nurses that will explore the effectiveness of realistic facial expressions in critical care settings, and may explore psychiatric care settings (Moosaei et al., 2014; Riek & Robinson, 2011).

**Diagnosing and Studying Schizophrenia**

Robotics technology has also been used in novel ways to study people with schizophrenia. Lavelle, Healey, and McCabe (2013, 2014) ran a series of experiments in which people with schizophrenia were placed in triads with matched control participants who were unaware of their diagnosis. Each group was tracked by a Vicon motion capture system, and their body motion was analyzed in terms of their social rapport and group participation. Interaction between 20 mixed triads was compared to 20 control triads (total $n = 120$). The results supported the hypothesis that people with schizophrenia truly do experience reduced rapport during first-time meetings when compared with controls.

The work by Lavelle et al. was the first to quantitatively explore the natural interaction between people with schizophrenia and matched controls on such a detailed level. Since this work has been published, new sensors have gone on the market that are a fraction of the cost of a Vicon motion capture system and provide even more detail. Thus, this technology offers researchers and practitioners an entirely novel way to study interpersonal interactions.

For example, our own recent work uses non-linear analysis techniques and motion modeling to explore how teams synchronize with one another during psychomotor entrainment tasks (Iqbal & Riek, 2014, 2015; Iqbal, Gonzales, and Riek, 2015; Rack, Iqbal, and Riek, 2015). As synchronizing with others is a key indicator of brain health and social development, the methods we are developing may be useful for therapists and researchers working with individuals with autism, post-traumatic stress disorder, and traumatic brain injury.

## Design Issues

## Potential Barriers to Provider Technology Adoption

Robot designers should be cognizant of the fact that mental healthcare professionals may be hesitant to embrace robotics technology in their practice for several reasons. First, it may interfere with, or be perceived by patients to interfere with, face-to-face communication between the provider and patient. This has been a growing problem in treatment and safety across other healthcare disciplines, and there is all the more reason to suspect this problem would be magnified for a profession centered on face-to-face interaction (cf. Levinson, Gorawara-Bhat, & Lamb, 2012; Meeks et al., 2014; Montague & Asan, 2014).

Second, providers may feel robotics technology could interfere with or hinder their practice, such as has been found in some studies with telepre- sence psychiatric treatment delivery systems (May et al., 2001). The mere presence of robotics technology itself could be perceived by providers as a threat to care delivery, and without long-term, thorough RCTs showing the effectiveness of the new technology, this effect may be worsened.

Finally, just as with any new technology introduced into a profession, many latent barriers to adoption exist. Several scholars have identified mod- els for predicting technology adoption among various professionals, including mental health practitioners, which may be useful to employ (c.f. Aarons, 2004; Chau & Hu, 2002). Robotics technology presents a unique adoption barrier, due to its sordid cultural history in science fiction (Riek, Adams, & Robinson, 2011). It is important that robots are well-designed to quickly convey their true capabilities to end-users, to alleviate any concerns.

## Cultural Barriers to Technology Adoption

Mental healthcare professionals considering the use of robotics technology in their practice should also be sensitive to how clients from a given culture might accept a robot. Many of the studies reported in the literature were conducted with Japanese populations, who may have more positive attitudes toward some robot morphologies compared to other cultures (Bartneck, 2008; Hornyak, 2006). Wang et al. found in a cross- cultural study between Chinese and American participants, that robots that respect cultural norms are more likely to be accepted (Wang, Rau, Evers, Robinson, & Hinds, 2010). Lee et al. found similar findings when comparing Korean and American participants (Lee, Sung, Sabanovic, & Han, 2012).

Some cultures may be more accepting of robotics in healthcare practice than one might expect. For example, while traditionally Middle Eastern culture is opposed to iconic technology, such as humanoid robots, people are often willing to make exceptions when it comes to health care. Riek et al. conducted a large-scale study ($n = 131$) in the United Arab Emirates, and brought an interactive android robot that resembled Ibn Sina to a shopping mall (see Figure 8.4; Riek et al., 2010). We surveyed participants from 21 different countries on their attitudes toward human-like robots, and found that generally participants were accepting of robots in domestic life (e.g., positive responses to discrete-visual analog scale questions such as "I wouldn't mind if a human-like robot treated me at a hospital," "I wouldn't mind if a human-like robot cleaned my house," etc.). More work is needed to fully understand the potential effect of culture on robotics technology adoption for mental healthcare use, but the robot's morphology and behavior are key components to consider.

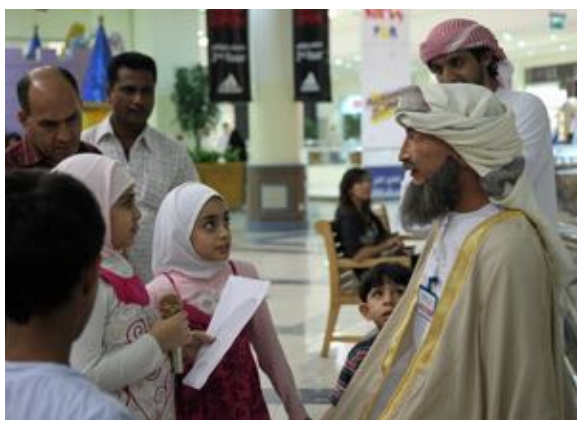

**Figure 8.4** Participants at a shopping mall in the United Arab Emirates interact with a humanoid robot resembling the philosopher Ibn Sina. Despite being situated in a culture that is frequently associated with anti-iconography, such a robot was found to be well-accepted if used for hospital care, and for domestic tasks such as housework (Riek et al., 2010).

## A Need for Evidence-Based Robotics Use in Mental Health Care

As discussed previously in this chapter, before robots can be effectively deployed in mental healthcare practice it is important that rigorous RCTs are conducted. These studies should study not only patient benefit and technology acceptance issues, but also explore the socio-technological needs of caregivers (Chang and Sábanović, 2015). While it is unlikely most robots would cause direct harm to clients or practitioners, adopting this technology before the evidence base is developed runs the risk of displacing proven interventions with less effective or ineffective treatments.

Thus, robot designers wishing to explore the use of robotics technology in mental health care should, from the outset, aim to run rigorous studies in concert with practitioners and researchers. This is particularly important for several reasons. First, it is important to ensure suitable controls are selected. For example, rather than the treatment group receiving the robot and the control group receiving an alternate therapy, the control group could instead receive a non-actuated analog. This strategy was successful for the aforementioned Kobachan intervention for age-related cognitive decline (Tanaka et al., 2012), and thus may prove useful for other interventions.

Second, by partnering with practitioners, robot designers are more likely to have access to desired patient populations, and are more likely to be able to engage in long-term studies. For many mental healthcare applications, clinical effectiveness is related to treatment length, i.e., the dose-response relationship (Hansen, Lambert, & Forman, 2002). Thus, a single encounter with a robot is not as useful a metric, particularly given the high likelihood of novelty effects (Gockley et al., 2005; Leite, Martinho, Pereira, & Paiva, 2009).

Finally, both robot designers and mental healthcare professionals should be sensitive to the fact that people with disabilities and their families can be swept up in treatment fads and pseudoscience, some of which can be truly damaging (Jacobson, Foxx, & Mulick, 2005). Despite the allure of new technology, it is better to tread carefully and run thorough RCTs before embracing new treatment modalities.

## Ethical Issues

Riek and Howard (2014) recently published a paper describing unique ethical challenges in HRI, and proposed a code of ethics for HRI practitioners. HRI practitioners include: robotics designers, engineers, researchers, designers, product managers, and marketers, and also may include mental healthcare professionals and other healthcare workers interested in exploring the use of robots in their practice. Riek and Howard (2014) specifically focus on ethical challenges to using robots with vulnerable populations, including socially and physically assistive robots.

The proposed code of ethics is recounted here verbatim (see Riek & Howard, 2014 for full details and examples of unique ethical challenges):

### The Prime Directive

All HRI research, development, and marketing should heed the overall principle of respect for human persons, including respect for human autonomy, respect for human bodily and mental integrity, and the affordance of all rights and protections ordinarily assumed in human-human interactions. The robot actor is expected to behave in a manner at least as respectful of human personhood as human actors to the extent feasible.

### Specific Principles

**Human Dignity Considerations**

- The emotional needs of humans are always to be respected.

- The human's right to privacy shall always be respected to the greatest extent consistent with reasonable design objectives.

- Human frailty is always to be respected, both physical and psychological.

**Design Considerations**

- Maximal, reasonable transparency in the programming of robotic systems is required.

- Predictability in robotic behavior is desirable.

- Trustworthy system design principles are required across all aspects of a robot's operation, for both hardware and software design, and for any data processing on or off the platform.

- Real-time status indicators should be provided to users to the greatest extent consistent with reasonable design objectives.

- Obvious opt-out mechanisms (kill switches) are required to the greatest extent consistent with reasonable design objectives.

**Legal Considerations**

- All relevant laws and regulations concerning individuals' rights and protections (e.g., FDA, HIPPA, and FTC) are to be respected.

- A robot's decision paths must be re-constructible for the purposes of litigation and dispute resolution.

- Human informed consent to HRI is to be facilitated to the greatest extent possible consistent with reasonable design objectives.

**Social Considerations**

- Wizard-of-Oz should be employed as judiciously and carefully as possible, and should aim to avoid Turing deceptions.

- The tendency for humans to form attachments to and anthropomorphize robots should be carefully considered during design.

- Humanoid morphology and functionality is permitted only to the extent necessary for the achievement of reasonable design objectives.

- Avoid racist, sexist, and ableist morphologies and behaviors in robot design.

Since this initial paper was published, we have run a workshop entitled "The Emerging Policy and Ethics of Human-Robot Interaction" (Riek et al., 2015). The workshop has further expanded upon these issues, and we will shortly release a consensus document to the community.

# CONCLUSION

The personal robotics sector is expanding rapidly, with new companies forming, products being developed, and applications appearing in the commercial space in ways unimaginable a decade ago. While it is impossible to form specific predictions, the current trend of research and development suggests a future where robots are able to assist us in a variety of daily tasks. For people with physical disabilities in particular, robots are uniquely poised to be a remarkably enabling technology, affording an improved quality of life and an increase in independent living.

Robots may also help individuals with invisible health conditions, such as mental health disorders, and they may also assist their caregivers. As one participant stated at our HRI Policy and Ethics Workshop (Riek et al., 2015), often people just want a "robot that can change the oil." In other words, a robot that can help with mundane chores around the house or daily living tasks. Simple things can prove hugely beneficial.

For care providers, robots may prove useful in training. For example, similar to how researchers in the virtual agent communities have employed both virtual patients and coaches to train clinical students (Lok et al., 2006; Rizzo, 2002), perhaps robots could also be used in this capacity. For example, to teach interaction, listening skills, and enable destigmatization (Arkin, 2014; Riek & Robinson, 2011).

However, providers are advised caution in using autonomous robots to directly provide therapy for individuals with mental health disorders. It is important that a strong evidence base is first established through the employment of rigorous RCTs. Despite the appeal of "Sigfrid von Shrink," the robotic psychologist in Fredrick Pohl's novel Gateway (Pohl, 1977), the actual realization of AI has nowhere near caught up to the remarkable physical capabilities of robots. That being said, this is indeed an exciting era for robotics technology. As computer processors and storage get faster and cheaper, and as new advances emerge in machine learning and multimodal processing, a world of possibility exists for robots ever more capable and agile in human social environments.

# ACKNOWLEDGMENT

Some research reported in this article is based upon work supported by the National Science Foundation under Grant No. IIS-1253935.

Hornyak, T. N. (2006). Loving the machine. Tokyo, Japan: Kodansha International.